\documentclass{article}
\usepackage[T5,T1]{fontenc}
\usepackage{microtype}
\usepackage{graphicx}
\usepackage{subfigure}
\usepackage{booktabs} 

\usepackage{hyperref}

\usepackage[accepted]{icml_codeml_2025}

\usepackage{amsmath}
\usepackage{amssymb}
\usepackage{mathtools}
\usepackage{amsthm}

\usepackage[capitalize,noabbrev]{cleveref}

\theoremstyle{plain}

\theoremstyle{definition}

\theoremstyle{remark}

\usepackage[disable,textsize=tiny]{todonotes}

\usepackage{comment}
\usepackage{xspace}
\usepackage{adjustbox}
\usepackage{listings}

\usepackage{xcolor}
\definecolor{codegreen}{rgb}{0,0.6,0}
\definecolor{codegray}{rgb}{0.5,0.5,0.5}
\definecolor{codepurple}{rgb}{0.58,0,0.82}
\definecolor{backcolour}{rgb}{0.95,0.95,0.92}
\usepackage[table]{xcolor}
\usepackage{pifont}
\usepackage{caption}

\definecolor{lightgreen}{RGB}{210, 250, 210}
\definecolor{darkgreen}{RGB}{0, 153, 0}
\definecolor{lightgrey}{RGB}{240, 240, 240}
\definecolor{lightblue}{RGB}{220, 235, 255}
\definecolor{darkblue}{RGB}{0, 70, 140}   
\newcommand{\tickcell}{\cellcolor{lightblue}\ding{51}}
\newcommand{\xcell}{\cellcolor{lightgrey}~}
\newcommand{\discotick}{\cellcolor{darkblue}

{\color{white}\ding{51}}}

\lstdefinestyle{mystyle}{
    language=Python,
    backgroundcolor=\color{backcolour},   
    commentstyle=\color{codegreen},
    keywordstyle=\color{magenta},
    numberstyle=\tiny\color{codegray},
    stringstyle=\color{codepurple},
    basicstyle=\ttfamily\footnotesize,
    breakatwhitespace=false,         
    breaklines=true,                 
    captionpos=b,                    
    keepspaces=true,                 
    numbers=left,                    
    numbersep=5pt,                  
    showspaces=false,                
    showstringspaces=false,
    showtabs=false,                  
    tabsize=2
}

\newcommand{\disco}{DISCO\xspace}
\newcommand{\discollaborative}{DISCOllaborative\xspace}
\newcommand{\discollaboratives}{DISCOllaboratives\xspace}

\usepackage{pifont}
\usepackage{hhline}
\icmltitlerunning{\disco: A Browser-Based Privacy-Preserving Framework for Distributed Collaborative Learning}

\begin{document}

\twocolumn[
\icmltitle{\disco: A Browser-Based Privacy-Preserving Framework\\ for Distributed Collaborative Learning}



\icmlsetsymbol{equal}{*}

\begin{icmlauthorlist}
\icmlauthor{Julien {\fontencoding{T5}\selectfont Tu\'\acircumflex{}n T\'u} Vignoud}{EPFL}
\icmlauthor{Valérian Rousset}{EPFL}
\icmlauthor{Hugo El Guedj}{EPFL}
\icmlauthor{Ignacio Aleman}{EPFL}
\icmlauthor{Walid Bennaceur}{EPFL}
\icmlauthor{Batuhan Faik Derinbay}{EPFL}
\icmlauthor{Eduard Ďurech}{EPFL}
\icmlauthor{Damien Gengler}{EPFL}
\icmlauthor{Lucas Giordano}{EPFL}
\icmlauthor{Felix Grimberg}{EPFL}
\icmlauthor{Franziska Lippoldt}{EPFL}
\icmlauthor{Christina Kopidaki}{EPFL}
\icmlauthor{Jiafan Liu}{EPFL}
\icmlauthor{Lauris Lopata}{EPFL}
\icmlauthor{Nathan Maire}{EPFL}
\icmlauthor{Paul Mansat}{EPFL}
\icmlauthor{Martin Milenkoski}{EPFL}
\icmlauthor{Emmanuel Omont}{EPFL}
\icmlauthor{Güneş Özgün}{EPFL}
\icmlauthor{Mina Petrović}{EPFL}
\icmlauthor{Francesco Posa}{EPFL}
\icmlauthor{Morgan Ridel}{EPFL}
\icmlauthor{Giorgio Savini}{EPFL}
\icmlauthor{Marcel Torne}{EPFL}
\icmlauthor{Lucas Trognon}{EPFL}
\icmlauthor{Alyssa Unell}{EPFL}
\icmlauthor{Olena Zavertiaieva}{EPFL}
\icmlauthor{Sai Praneeth Karimireddy}{EPFL}
\icmlauthor{Tahseen Rabbani}{uot}
\icmlauthor{Mary-Anne Hartley}{EPFL}
\icmlauthor{Martin Jaggi}{EPFL}

\end{icmlauthorlist}

\icmlaffiliation{EPFL}{School of Computer and Communication Sciences, EPFL (Ecole polytechnique fédérale de Lausanne), Lausanne, Switzerland}
\icmlaffiliation{uot}{University of Chicago, USA}

\icmlcorrespondingauthor{Julien Vignoud}{julien.vignoud@epfl.ch}
\icmlcorrespondingauthor{Martin Jaggi}{martin.jaggi@epfl.ch}

\icmlkeywords{Federated Learning, Decentralized Learning, Privacy-preserving, Collaborative Learning}

\vskip 0.3in
]



\printAffiliationsAndNotice{}  
\begin{abstract}
Data is often impractical to share for a range of well considered reasons, such as concerns over privacy, intellectual property, and legal constraints. This not only fragments the statistical power of predictive models, but creates an accessibility bias, where accuracy becomes inequitably distributed to those who have the resources to overcome these concerns. We present \textbf{\disco}\footnote{Code repository is available at \url{https://github.com/epfml/disco/} and a showcase web interface at \url{https://discolab.ai/}}: an open-source \textbf{Dis}tributed \textbf{Co}llaborative learning platform accessible to non-technical users, offering a means to collaboratively build machine learning models without sharing any original data or requiring any programming knowledge. DISCO's web application trains models locally directly in the browser, making our tool cross-platform out-of-the-box, including smartphones. The modular design of \disco offers choices between federated and decentralized paradigms, various levels of privacy guarantees and several approaches to weight aggregation strategies that allow for model personalization and bias resilience in the collaborative training.
\end{abstract}

\section{Introduction}
\label{sec:intro}

By 2025, global data is estimated to reach 168 zettabytes, with a projected 25 petabytes of genomic data \citep{Banks2020-bigdata} and 50 petabytes of hospital data \citep{organization2019global} added annually. As the scale of big data increases and its granularity deepens, so too does its potential power, value, risk and the legal constraints of sharing it \citep{Horvitz2015-privacy}. Beyond privacy concerns, sharing data also involves issues of ownership, sovereignty, intellectual property, and fairness \citep{Bellantyne2020-ownership}. 

However, when data is not shared, the problems encountered are equally challenging. Its statistical power becomes fragmented, risking poor generalization \citep{zhao2018federated}. This disproportionately affects lower-resource sites that lack the capacity to ensure robust data collection or collaborative data security \citep{Zelch2018-generalization}. Protectively siloing data also creates interoperability drift and a burdensome harmonization task, further disincentivizing collaboration \citep{Crowson2022-interoperability}.

\begin{table*}[t!]
    \captionsetup{justification=centering}
    \caption{Feature comparison of existing collaborative training platforms. \\\textsuperscript{1}{DP = Differential Privacy.} \textsuperscript{2}{Supports Decentralized Learning.}}
\centering
\arrayrulecolor{white}
\setlength{\arrayrulewidth}{1pt}
\small
\begin{tabular}{r|c|c|c|c|c|c}
    & DP\footnotemark & Encryption & Open Src & Browser client & Decentralized\footnotemark & Code-free training\\
    \hline
    SensiX \cite{min2020sensix}                 & \xcell & \xcell & \xcell & \xcell & \xcell & \xcell \\
    \hline
    HomoPAI \cite{li2020homopai}                 & \xcell & \tickcell & \xcell & \xcell & \xcell & \xcell \\
    \hline
    IntegrateAI \cite{Crowson2022-interoperability}            & \tickcell & \tickcell & \xcell & \xcell & \xcell & \xcell \\
    \hline
    Nvidia FLARE \cite{roth2022nvidia}            & \tickcell & \tickcell & \tickcell & \xcell & \xcell & \xcell \\
    \hline
    Flower \cite{beutel2020flower}                 & \tickcell & \tickcell & \tickcell & \xcell & \xcell & \xcell \\
    \hline
    FATE \cite{liu2021fate}                   & \xcell & \tickcell & \tickcell & \xcell & \xcell & \xcell \\
    \hline
    PySyft \cite{ziller2021pysyft}                & \tickcell & \tickcell & \tickcell & \xcell & \xcell & \xcell \\
    \hline
    TF Federated \cite{mcmahan2017communication}  & \tickcell & \tickcell & \tickcell & \xcell & \xcell & \xcell \\
    \hline
    \textbf{DISCO} (this work)        & \discotick & \discotick & \discotick & \discotick  \textcolor{white}{\textit{ (incl. mobile)}} & \discotick & \discotick \\
\end{tabular}
    \label{tab:platform_comparison}

\end{table*}

Indeed, secure centralization of the data requires significant resources, and laborious applications for ethical approval, which are not granted in perpetuity, and thus do not align with the nature of the exponentially amassing live streams of big data.

To fully harness the potential of massively distributed data, the learning process can be similarly distributed to the data sources. Collaborative learning thus involves exchanging updates from local learning steps, typically in the form of gradients, which are then aggregated into a common model. In federated learning (FL), the gradient aggregation process is coordinated by a central server, whereas in decentralized learning (DL), it is delegated to individual nodes. While the concept of parallel computing is not new, the term 'federated learning' was formalized in 2016 \citep{konevcny2016federated, mcmahan2017communication}. It is understood as a set of challenges faced when deriving machine learning models from multiple nodes \citep{kairouz2021advances}.

Although collaborative learning appears to address most challenges of data sharing, several potential issues and vulnerabilities can hinder its practical implementation and scalability. For instance, while FL addresses the issue of data sovereignty and ownership, it places the intellectual property of the model and gradients at the discretion of a central entity. This central server also potentially represents a single point of failure, which can become a bottleneck in high-communication settings \citep{lian2017decentralized}. While DL can resolve this issue, it comes with its own constraints such as limited computational power in end-nodes (e.g., smartphones). 

Neither FL nor DL are immune to attacks on privacy \citep{Mothukori-privacy, Troncoso-privacy, geiping2020inverting,wen2022fishing} and require a battery of features to stave off threats to GDPR compliance \citep{Truong2021-FL-GDPR}. Various kinds of privacy-enhancing techniques can be combined with collaborative training, including differential privacy \citep{abadi2016deep, wei2020federated}, secure multiparty computation \citep{bonawitz2017practical,mohassel2017secureml, smpc2024}, and homomorphic encryption \citep{zhang2020batchcrypt}, etc. 

For collaborative learning to be practically feasible and realize its potential to improve fairness, representation, and accessibility, an effort must be made to create a customizable suite of easy-to-use, open-source resources. In this work, we propose \disco: an open-access, open-source \textbf{Dis}tributed \textbf{Co}llaborative learning platform, offering users a means to collaboratively build machine learning models without sharing any original data. The modular design of \disco allows users to leverage a flexible combination of features according to their requirements. This includes choices regarding federated and decentralized communication protocols, various levels of privacy guarantees, and several approaches to gradient aggregation strategies. 

While other FL platform implementations exist, none support DL or offer the same level of accessibility to non-technical users that \disco provides through its code-free, installation-free client. We show a comparison of existing frameworks in Table~\ref{tab:platform_comparison} and describe them individually in Appendix~\ref{sec:relatedwork}. The main contributions of \disco are:
\begin{itemize}
    \item An in-browser and code-free web application supporting local deep learning model training, including on smartphones. Its intuitive user interface eliminates the need for data owners to possess technical expertise required to install software and implement collaborative learning scripts, a common requirement for other platforms.
    \item Support for decentralized learning, enabling participants to train models collaboratively without sharing weight updates with a central server. Furthermore, \disco can leverage secure multi-party computation to enhance model privacy against other participants.
\end{itemize}
\section{The \disco library}
\label{sec:disco}
\disco is a novel open-source software for collaborative machine learning. It is the first in-browser collaborative training platform supporting both federated and decentralized learning, and provides a code-free interface allowing users to create and join collaborative machine learning tasks, termed "\textbf{\discollaboratives}" and are described in Section~\ref{sec:discollaboratives}. \disco enables participants to jointly train deep learning models without sacrificing data ownership by sharing model weight updates — rather than raw data — through collaborative learning. It also provides users with a number of options to accommodate their specific training and privacy requirements.

DISCO is designed for two primary use cases. The first is \textit{public}, large-scale collaborative training, where \discollaboratives are open to anyone to join by contributing readily accessible data, and new \discollaboratives can be created by any user. For instance, users can leverage \disco to train spam detection systems by contributing their personal spam emails, speech recognition systems by contributing voice recordings based on provided transcripts or for image classification by labeling medical or non-medical images. A public \disco instance\footnote{\url{https://discolab.ai/}} already showcases this use case.

The second use case envisioned is a \textit{private} intra-organizational training, for instance, within a network of hospitals—where the generalization of medical models would significantly benefit from inter-hospital collaboration, but hindered due to patient data confidentiality and data sharing laws. In such scenarios, training is restricted to selected entities aiming to train models collaboratively without direct data sharing, typically deploying a custom private \disco instance with whitelisted access.

Consequently, \disco is not a single website; numerous instances can be deployed, both public and private, that accommodate for its diverse use cases.

\subsection{Architecture}

\disco features a modular architecture comprising several key components: a core library complemented by 2 platform-specific packages, a server for communication orchestration, and two distinct user interfaces - a web application and a command line interface. This design promotes reusability and composability. To enable in-browser operation and ensure that \disco remains easy to customize and adapt, all modules - including the machine learning stack and backend components - are implemented in JavaScript/TypeScript. Each of these modules is also open-source, allowing users to further customize \disco according to their specific needs.

\textbf{Core library.} The \texttt{discojs} core library is platform agnostic, designed to run in any JavaScript/TypeScript engine. It forms the foundational codebase of \disco, containing type definitions utilized throughout the system. Key components include a \texttt{Dataset} class for efficient data loading and composition, a \texttt{Trainer} to manage local model training synchronized with server updates, various \texttt{Aggregator} implementations offering different levels of security, and a collection of \texttt{Model}. The library's generic design permits the use of any framework for model definition; currently, models are implemented using the TensorFlow.js library \citep{smilkov2019tensorflow}.

Two platform-specific packages complement the library for browser and Node.js environments (respectively \texttt{discojs-web} and \texttt{discojs-node}), providing helper functions that facilitate its use and leverage platform-specific technologies, such as a compiled TensorFlow library on Node.js and \texttt{Canvas} rendering in the browser.

\textbf{Clients.} The web application is \disco's code-free user interface, through which users join \discollaboratives. All computation occurs directly within the browser, with no off-loading to local programs. This design ensures broad accessibility: no installation is required, and the platform can even run on mobile devices. The web application primarily targets non-technical users; no domain-specific knowledge is required and users can join training sessions by connecting their data and initiating the process. The interface also supports the creation of new \discollaboratives and using trained models for inference.

As an alternative to the web interface, \disco provides a Command Line Interface (CLI) relying on Node.js, enabling technical users to interact with \disco headlessly and facilitating large-scale experiments. Both CLI and browser-based users can participate in the same training session.

\textbf{Server.} The server, deployed by technical maintainers, hosts available \discollaboratives, orchestrates training sessions, and handles federated weight aggregation. In DL, it is also utilized for peer discovery and the pacing of training rounds.

\subsection{\discollaboratives: collaborative training tasks}
\label{sec:discollaboratives}
\textbf{Creation.} \discollaboratives can be set up via the code-free UI or programmatically. Each task includes a description of the training goal, the base model architecture (currently in TensorFlow.js format), the expected format of data, and customizable hyperparameter settings. These hyperparameters include epochs, batch size, differential privacy settings among others. DISCO's public instance showcases several pre-defined tasks. These \discollaboratives include classical LLM training and MNIST classification or COVID-19 diagnosis from lung ultrasounds.

\textbf{Training.} Participating users are first shown a description of the \discollaborative and prompted to connect the data they wish to use for training. Importantly, \textbf{connected data is not uploaded anywhere and remains on the local device}, the UI reads this data solely for local model training. After connecting their data, users can directly join the collaborative training session. Note that users can also decide to train locally on their own. The training is paused if there are insufficient participants in the session. 

\begin{figure}[ht]
    \centering
    \includegraphics[width=\columnwidth]{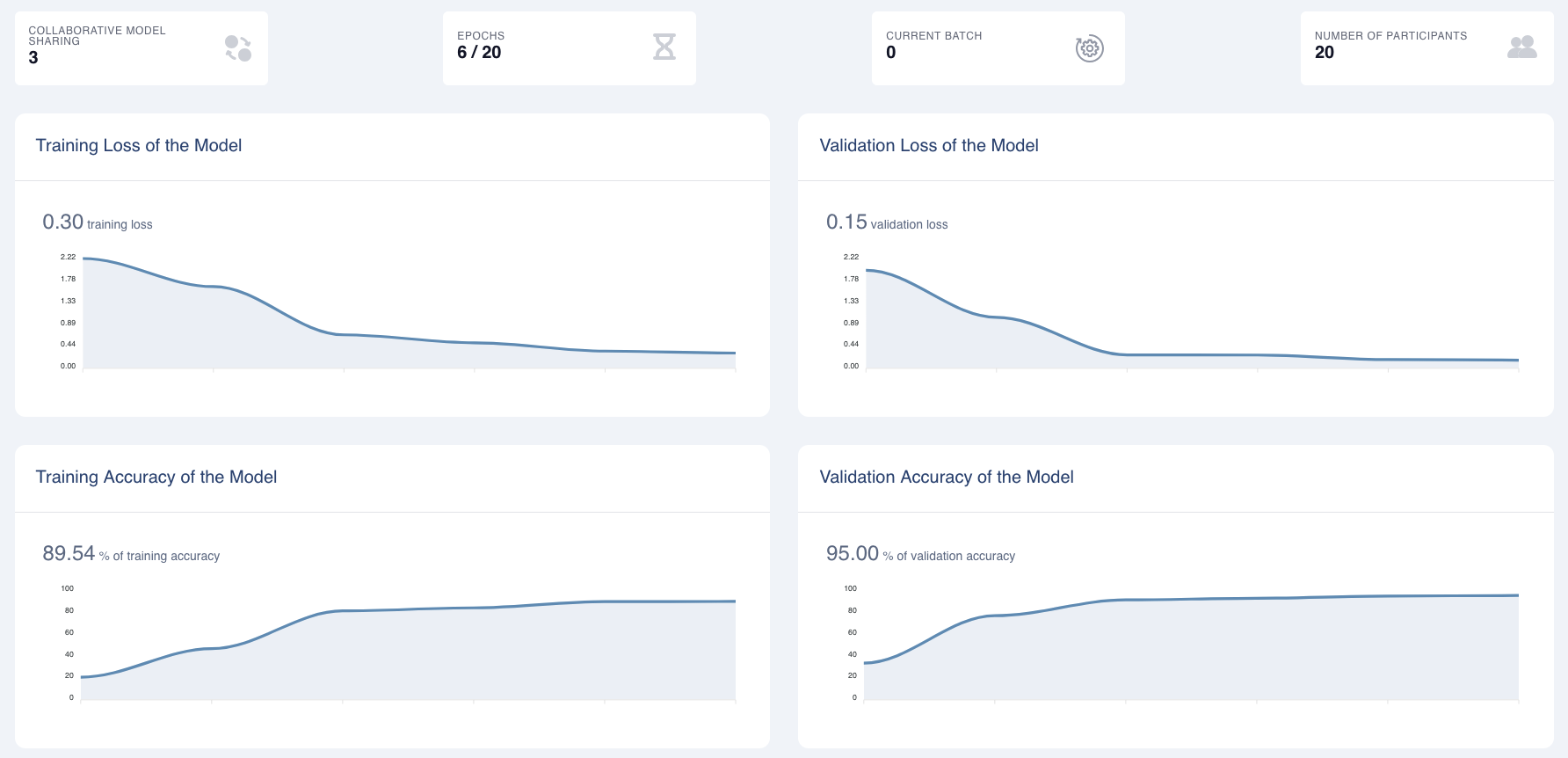}
    \vskip -0.1in
    \caption{Performance visualization for a federated MNIST classification task.}
    \label{fig:performance}
    \vskip -0.1in
\end{figure}

Any user participating in a task can monitor their local model's performance and view information regarding the number of actively participating users in the session. Figure~\ref{fig:performance} depicts the training and test curves for a model that is globally aggregated every two epochs (over a total of 20 epochs). Note that these accuracies reflect performance on the individual client's data; consequently, these graphs will differ from client to client. After training, users can export the resulting model or utilize it within the \disco platform.

\subsection{In-browser Deep Learning}
Performing deep learning operations directly within the user's web browser is foundational to DISCO's accessibility. This is primarily achieved through TensorFlow.js \citep{smilkov2019tensorflow}, an open-source machine learning library maintained by Google specifically developed for JavaScript environments. TensorFlow.js supports multiple computational backends that implement tensor storage and mathematical operations. The WebGL backend is the most powerful one and leverages the system's GPU when available for hardware acceleration. With WebGL, tensors are stored as textures, and mathematical operations are implemented as WebGL shaders. TensorFlow.js can fall back on the CPU or WebAssembly backends. When using DISCO's CLI rather than the web application, TensorFlow.js can leverage its Node.js backend for hardware acceleration.

TensorFlow.js natively supports tabular data and images. \disco extends its capabilities to language modeling tasks by implementing the Transformers block in Tensorflow.js, resulting in a custom GPT2 implementation. Tokenization is performed using the Transformers.js library \citep{lochner2023transformers}. 

Transformers.js - and ONNX Runtime Web \cite{onnxruntime} on which it relies - are promising frameworks for browser-based ML, notably due to early WebGPU support (the successor to WebGL) and the possibility to convert existing pre-trained language models to Transformer.js. However, in-browser training is not yet supported by either libraries and to this date TensorFlow.js remains the most popular library implementing in-browser automatic differentiation for back-propagation.

\subsection{Privacy-preserving training}
\label{sec:privacy}
To implement differential privacy, task creation options enables adding Gaussian noise \citep{mcmahan2018learning} to gradient updates by setting a noise scale, as well as clipping gradients \citep{abadi2016deep} by adjusting a clipping radius. These measures aim to limit malicious or honest-but-curious reconstruction of cross-client training data statistics.

DL also supports secure multi-party computation (SMPC) \citep{smpc2024} for aggregating weight updates, thereby keeping them private from other participants. SMPC relies on additive secret sharing, wherein a secret - here a local weight update - is divided into shares. Each share is generated by adding a random value to a specific portion of the original secret. The shares are distributed among the parties, and the secret can only be reconstructed by combining a sufficient number of shares.
\todo[author=julien]{Is it useful to mention this?}  Furthermore, support for Byzantine-robust aggregation methods \citep{karimireddy2021learning} is currently under development.

\subsection{Collaborative Learning in \disco}

\textbf{Federated learning.} \disco relies on a central server to manage the aggregation of model updates and the distribution of the updated model architecture in each round. After training the model locally in the web application, each client uploads its model update (as a gradient) to the server.  The server aggregates all updates using the classical \textsc{FedAvg} algorithm \citep{mcmahan2017communication} and sends a global update back to all clients. \todo[author=julien]{Is it useful to mention this?} Implementing robust aggregation methods, such as \textsc{FedProx} \citep{li2020fedprox} is currently underway.  \disco clients establish WebSocket connections to communicate with the server, allowing the latter to be notified when clients leave the training session.

\textbf{Decentralized learning.} In contrast, nodes in DL do not share weights with a central server; instead, they exchange them directly, a process that occurs within the clients' web browsers. Specifically, \disco nodes establish peer-to-peer connections using WebRTC (Web Real-Time Communication) which is the only way to ensure direct connections between peers within the browser environment. WebRTC's capabilities, such as NAT traversal (including hole punching for firewalls) and support for re-establishing dropped connections, allow \disco to operate across diverse network configurations. 

Although the server never accesses weight updates in DL, it remains essential for orchestrating the training session. This orchestration includes assigning client IDs, designating peers for participation in weight-sharing rounds, and acting as a WebRTC \textit{signaling server} to facilitate peer discovery and connection establishment.

At each weight sharing round, nodes first notify the server of their intent to join. Each node then trains locally for a predetermined number of epochs before informing the server of its readiness to exchange weights. Once the number of peers ready to exchange weights reaches a threshold defined during task creation, the server shares the list of these peers with the nodes participating in the round. Peers then establish peer-to-peer connections to exchange weight updates, forming a fully connected topology. Weight updates can then be aggregated either by simple averaging or by leveraging secure multi-party computation as described in Section~\ref{sec:privacy}. 

The availability of decentralized training is a unique feature of \disco, not currently supported by comparable collaborative learning platforms (see Table~\ref{tab:platform_comparison}).

\section{Conclusion}

We introduced \disco, a novel open-source platform for collaborative machine learning that preserves data sovereignty. By supporting both federated and decentralized learning, and integrating flexible privacy safeguards, \disco enables model training without the need to centralize raw data. Its no-install, no-code interface—including support for mobile browsers—lowers technical barriers to entry, fostering broader participation in AI development. In doing so, \disco contributes to more inclusive, equitable, and privacy-preserving machine learning ecosystems.

\section*{Impact Statement}

This paper presents work whose goal is to advance the field of 
Machine Learning. There are many potential societal consequences 
of our work, none of which we feel must be specifically highlighted here.

\bibliography{references}
\bibliographystyle{icml2025}

\newpage
\appendix
\onecolumn

\section{Related work}
\label{sec:relatedwork}

\paragraph{SensiX}\citep{min2020sensix} and its successor SensiX++ \citep{min2023sensix++} are multi-device, multi-modal runtimes tailored for edge-based collaborative learning. Clients served by this system component include sensory devices such as earbuds, smartwatches and smartphones. SensiX lies between the sensors and sensory model(s), adaptively pooling data from subsets of clients depending on data quality, energy costs, device availability, and desired task. SensiX is demonstrated for a variety of sensory tasks such as human activity recognition and audio keyword spotting, but only in a setup consisting of trusted wireless devices. In particular, privacy is not considered: the centralized SensiX controller is given unrestricted access to sensor data. Furthermore, as opposed to federated learning, the sensory model outputs do not improve the on-device models of the sensors. The SensiX library is not accessible to the public. 

\paragraph{HomoPAI} \citep{li2020homopai} is a secure collaborative protocol built on top of Alibaba Cloud's Platform for AI (PAI). The system enables multiple parties to train a central via model through submission of encrypted local data. Homomorphic encryption (HE) is performed on both user data and the model to enable training securely -- only the final model is decrypted and sent out to users. Peer-to-peer (i.e., decentralized) learning is also available. HomoPAI also makes use of MPI to conduct training in parallel. Similar to FL, the protocol attempts to prevent exposure of raw client data. However, in HomoPAI, data is submitted to the central server, as opposed FL, where only gradients are uploaded. A demo is provided for logistic regression, though deep neural networks are untested (with only pseudo-homomorphic approximations of activation functions such as the ReLU. It is stated that accuracies are suboptimal when compared to training over raw data, whereas modern FL matches single-machine training over a wide cluster of tasks. Furthermore, conducting operations over HE data, especially at the scale of a deep network is computationally expensive. The library is not open-source and it is unclear if it still being maintained. 

\paragraph{integrateAI} (\url{https://www.integrate.ai/}) is an online federated learning platform. The motivation is to enable pools of trusted collaborators to jointly train a model according to the principles of centralized FL. The motivation and design of integrateAI is quite similar to \disco, with features such as DP gradients, cloud-based or local training, and custom dataset loading. However, unlike \disco, the integrateAI platform is closed-source (API-based), doesn't work on mobiles or browser, and does not support decentralized learning.

\paragraph{NVIDIA FLARE} (NVIDIA Federated Learning Application Runtime Environment) \citep{roth2022nvidia} is a domain-agnostic, open-source federated SDK. FLARE supports various aggregation functions (FedAvg, FedProx), custom datasets, arbitrary architectures, and privacy-preserving model updates via various DP mechanisms and homomorphic encryption (HE). However, FLARE does not support decentralized learning and as an SDK, it is intended for application development and requires engineered setup, as opposed to than drop-in browser-based deployment like \disco.

\paragraph{Flower} \citep{beutel2020flower} Flower is a federated learning programming library that provides a unified approach to federated learning by remaining agnostic machine learning frameworks, meaning it is compatible with popular frameworks such as PyTorch, TensorFlow, Keras, scikit-learn, and Hugging Face. It is also platform-independent, allowing interoperability across different operating systems and hardware, which is beneficial for heterogeneous edge device environments. Flower emphasizes usability, enabling users to create a complete federated learning system with a small amount of Python code.

\paragraph{FATE} (Federated AI Technology Enabler) \citep{yang2021industrial} is focused on the industry-scale federated learning. FATE supports horizontal and vertical FL, linear/deep/tree-based models, and transfer learning. For gradient-boosted tree models, FATE supports the SecureBoost \cite{cheng2021secureboost} algorithm for enhanced privacy, along with model-agnostic approaches such as secure multi-party computation (MPC), HE, and general DP mechanisms. FATE-Flow is a multi-party federated task security scheduling platform which includes resource coordination, real-time job monitoring, multi-party cooperation authority management, CLI and Restful APIs. FATE does not support decentralized learning nor have a browser platform. 

\paragraph{PySyft}\citep{ziller2021pysyft} is a large open-source library for federated learning.
Data may be held locally or distributed by a central server across workers (for an internal and/or trusted network). PySyft enables both encrypted training and differentially private/secure gradient exchange. The platform relies on a "datasite" which is a centralized server clients may log into. Datasites may be launched using Docker or Kubernetes. A defining feature of PySyft is its interoperability: it supports PyTorch and TensorFlow training and mobile-friendly programming languages such as Kotlin and Swift. Decentralized learning is not supported as only a single node can send back results, and while it has a JavaScript version, \texttt{syft.js}, it hasn't been updated in years so currently it does not support a browser application. 

\paragraph{TensorFlow Federated (TFF)} (\url{https://www.tensorflow.org/federated}) is an open-source framework within the TensorFlow ecosystem for centralized federated learning. TFF is organized into a high-level layer for general implementation of federated training to existing models and a low-level layer for the construction of communication operators, aggregation functions, and foundational algorithms. Support for differential privacy is provided, and parts to build secure aggregation. Furthermore, update and model compression is supported for low-resource systems. Decentralized learning is unavailable, and the framework is intended for integrated engineering for existing TensorFlow routines as opposed to drop-train training for edge clients. It is solely in Python thus can't be run in a browser.


\end{document}